\documentclass[letterpaper,10pt,journal]{IEEEtran}

\usepackage{amsmath,amssymb}
\usepackage{graphicx}
\usepackage{booktabs}
\usepackage{multirow}
\usepackage{cite}
\usepackage{url}
\usepackage{xcolor}

\graphicspath{{figures/}}
\newcommand{\R}{\mathbb{R}}
\newcommand{\SO}{\mathrm{SO}}
\newcommand{\relu}{\operatorname{ReLU}}
\newcommand{\veeop}{\operatorname{vee}}
\newcommand{\p}{\mathbf{p}}
\newcommand{\vct}{\mathbf{v}}
\newcommand{\avec}{\mathbf{a}}
\newcommand{\jvec}{\mathbf{j}}
\newcommand{\w}{\boldsymbol{\omega}}
\newcommand{\xijet}{\boldsymbol{\xi}}

\newtheorem{assumption}{Assumption}
\newtheorem{proposition}{Proposition}
\newtheorem{theorem}{Theorem}
\newtheorem{corollary}{Corollary}

\title{Neural Differential Flatness with Learned Inverse Mappings for Feasible Trajectory Optimization
\author{Luyuhang~Duan, Can~Cui, Yan~Zhang, Li~Fan, and Chao~Xu}
\thanks{This research is funded by the "Pioneering and Leading Goose + X" Science and Technology Program of Zhejiang, Grant No.2025C01052. (Corresponding authors: Li~Fan; Chao~Xu)}
\thanks{Luyuhang~Duan is with the College of Control Science and Engineering, Zhejiang University, Hangzhou 310027, China (e-mail: 12232048@zju.edu.cn).}
\thanks{Can~Cui is with Huzhou Institute, Zhejiang University, Huzhou 313000, China (e-mail: sunnyeternally@163.com).}
\thanks{Yan~Zhang is with Huzhou Institute, Zhejiang University, Huzhou 313000, China (e-mail: 1871079816@qq.com).}
\thanks{Li Fan is with Huzhou Institute, Zhejiang University, Huzhou 313000, China, and also with Zhejiang University, Hangzhou 310027, China (e-mail:
	fanli77@zju.edu.cn).}
\thanks{Chao Xu is with the Institute of Cyber-Systems and Control, College of Control Science and Engineering, Zhejiang University, Hangzhou 310027, China,
	and also with Huzhou Institute, Zhejiang University, Huzhou 313000, China (e-mail: cxu@zju.edu.cn).}}

\begin{document}
\maketitle

\begin{abstract}
Model-based motion planning methods such as model predictive control are standard for
robots, but often occupy considerable computational resources.
Differential flatness reduces this burden by expressing states and inputs
algebraically from a flat output and its derivatives. However, its inverse map
remains system specific and model dependent. We present neural differential
flatness (NDF), which represents a learned inverse map as a differentiable
neural decoder from a flat-output jet to planning-relevant
states and inputs. We stabilize decoder-in-the-loop gradient-based trajectory
optimization using derivative-aware heads, dynamics-consistency regularization,
and a training-support penalty. Once trained, the
decoder is fixed within a spline optimizer, allowing objectives on states and
inputs to act directly on trajectory parameters. When map and first-derivative
errors are uniformly bounded over the planning region, we bound reconstructed
dynamics errors and establish constraint preservation and convergence toward
exact-map plans. Experiments on Ackermann and tail-sitter systems demonstrate
reliable unseen-task planning, reduced online workload, and adaptation to a
tested actuator mismatch.
\end{abstract}


\section{Introduction}
\label{sec:introduction}

\IEEEPARstart{M}{odel-based} motion planning and optimal control remain
standard tools across unmanned robotic systems
\cite{mayne2000constrained,mellinger2011minimum,han2022efficient,quan2020survey},
whose performance depends on a prediction model obtained from physics or system
identification \cite{ljung1999system}. A model that omits relevant dynamics
produces unreliable predictions, while a highly detailed nonlinear model is harder to
identify and makes repeated online Model Predictive Control optimization more expensive
\cite{picotti2023gokart,salzmann2023realtime}.

Differential flatness reduces this online burden. For a flat system, all states
and inputs are algebraic functions of a flat output jet and finitely many of
its derivatives \cite{fliess1995flatness}, so planning can optimize the
flat-output trajectory without repeatedly integrating the dynamics
\cite{richter2016polynomial,tal2023aerobatic}. Though the differential flatness
framework largely decreases the online computational burden, an accurate
inverse map from the jet to system states and inputs must still be derived.
Efficient polynomial representations have made this principle practical for
large constrained problems. MINCO supports geometrically constrained
spatiotemporal optimization for multicopters \cite{wang2022geometrically}, and
a related joint spatial-temporal formulation has enabled autonomous micro-UAV
swarms to navigate cluttered natural environments \cite{zhou2022swarm}.
Constructing this map can require substantial domain-specific modeling,
symbolic derivation, and parameter identification, particularly for strongly
coupled dynamics or systems spanning multiple operating regimes. Errors in the
model structure or parameters then bias the reconstructed motion.

One data-driven alternative is to learn a forward or latent dynamics model and
use it as the prediction model in MPC
\cite{hewing2020cautious,picotti2023gokart,salzmann2023realtime}, which
captures dynamics that are missing from a nominal physical model. However,
every MPC update must still roll the learned model across the prediction
horizon and differentiate through that rollout.
Meanwhile, prediction errors may also accumulate over the horizon. Forward
model learning can therefore improve prediction fidelity, but it retains the
online optimization structure of MPC where computational burden remains the
intractable problem.

\begin{figure*}[t]
    \centering
    \includegraphics[width=0.98\textwidth,height=0.34\textheight,
    keepaspectratio]{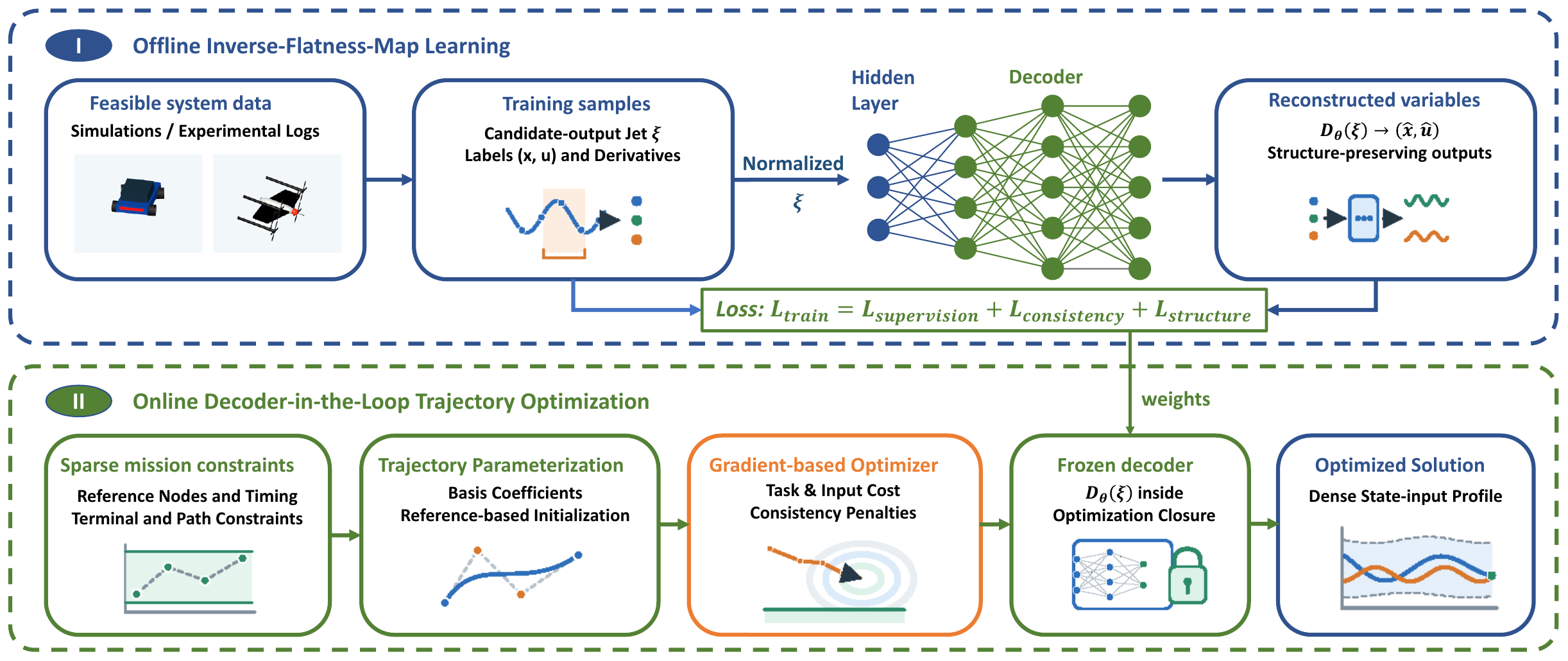}
    \caption{Method overview. A structured neural decoder is trained from
    feasible flat-output jets and physical labels. Supervised and
    dynamics-consistency objectives are used during training. At planning
    time, the decoder is frozen but remains in the automatic-differentiation
    graph. Sparse mission constraints then generate a new spline. The objective
    penalizes decoded actuation, physics residual, and empirical support.
    Returned trajectories are verified on a denser temporal grid.}
    \label{fig:overview}
\end{figure*}

We instead learn the inverse flatness map, where the candidate flat output jet has been specified. Feasible trajectories from simulations and real-world experiments then provide synchronized state
and input labels. The output and its derivatives form an interpretable
\emph{flatness layer}, then a structured neural decoder maps this layer to
planning-relevant states, controls, and auxiliary quantities. This design keeps
the optimization variables low-dimensional and physically meaningful, while
replacing a vehicle-specific symbolic inverse derivation with a differentiable
map that can adapt from deployment logs. Figure~\ref{fig:overview} summarizes the pipeline.

At planning time, the decoder weights
are fixed and the decoder is embedded in a spline optimizer. State and control
objectives then act directly on the flat-output parameters without recursive
state rollout. A support penalty discourages the optimizer from leaving the
region covered by training data. The returned trajectory is checked on a dense
temporal grid before acceptance.

The main contributions are as follows.
\begin{itemize}
    \item We develop a structured and differentiable inverse-flatness
    representation for a prescribed physical output, using continuous
    system-specific outputs and derivative-ordered decoder heads.
    \item We formulate a dynamics-consistency training objective and derive
    conditional bounds that connect inverse-map value and derivative errors to
    trajectory consistency, constraint margins, and convergence of planning
    solutions.
    \item We implement a support-aware spline planner that embeds the fixed
    decoder directly in trajectory optimization, with an analytical inverse map
    serving as the controlled baseline.
    \item We conduct trajectory-grouped, repeated-run, and ablation studies on
    an Ackermann vehicle and a full-envelope tail-sitter, evaluating unseen
    planning tasks, MPC benchmarks, and model-mismatch stress tests.
\end{itemize}

\section{Related Works}
\label{sec:related}

\subsection{Analytical Flatness-Based Planning}

Differential flatness characterizes systems whose states and inputs admit an
algebraic parameterization by a flat output and a finite derivative jet
\cite{fliess1995flatness}. Broad surveys of UAV motion planning distinguish
front-end path search from back-end continuous trajectory optimization
\cite{quan2020survey}. Within the latter, flatness parameterizations underlie
minimum-snap and polynomial trajectory generation
\cite{mellinger2011minimum,richter2016polynomial}. Wang \emph{et al.} develop
the MINCO representation to perform efficient spatial-temporal deformation
under geometric and dynamic constraints \cite{wang2022geometrically}. Zhou
\emph{et al.} integrate this optimization principle with onboard perception,
localization, and coordination for palm-sized UAV swarms operating in unknown
cluttered environments \cite{zhou2022swarm}. Other applications include
efficient spatiotemporal planning for ground vehicles \cite{han2022efficient}
and flatness-based
anti-sloshing trajectory generation with observer-assisted tracking
\cite{viet2024antislossing}. Flatness also supports aggressive aerial motion. Tal
\emph{et al.} use flatness for fixed-wing VTOL aerobatics
\cite{tal2023aerobatic}, while Lu \emph{et al.} derive and experimentally
validate a full-envelope tail-sitter construction
\cite{lu2024aggressive,lu2025autonomous}. These works establish the value of
flat-output planning, but their inverse maps remain tied to a system model,
its parameters, and a valid analytical chart. The present work retains the
same low-dimensional planning variables while replacing this system-specific
inverse computation on an audited domain.

\subsection{Learned Models in Trajectory Optimization}

Data-driven planning commonly learns forward evolution. Examples include
differentiable neural MPC \cite{salzmann2023realtime}, Gaussian-process
residual dynamics for cautious racing MPC \cite{hewing2020cautious}, and
continuous black-box acceleration models for real-time go-kart NMPC
\cite{picotti2023gokart}. These methods can model effects omitted by nominal
physics, but trajectory evaluation still requires horizon-wise rollout and
sensitivity propagation at each online update. Their errors can also
accumulate through repeated prediction. Complementary
robustness-oriented methods estimate disturbances and feed reachability
information back to the planner, as demonstrated for quadrotors subject to
motor damage and suspended-payload disturbances
\cite{zhou2024disturbanceaware}. Such methods improve resilience but retain an
explicitly designed forward-model and controller pipeline. NDF learns a
different object—a static inverse parameterization from the current
flat-output jet to the corresponding state and input. It therefore preserves
flat-output-space optimization without recursive state prediction.

Policy surrogates instead approximate the optimizer. Zhang \emph{et al.}
learn certified primal and dual MPC policies with backup control
\cite{zhang2021rapid}. NDF learns a plant-side inverse map and keeps mission
optimization explicit as objectives and constraints change.

\subsection{Learning within Flatness Structures}

Learning can enter a flatness-based pipeline at the output, inverse-map,
constraint, or feedback stage. Ma \emph{et al.} identify local flat outputs
from measured responses \cite{ma2022identification}. Their method addresses
which coordinates should be treated as flat, whereas NDF begins after a
physical output has been prescribed and learns how its jet reconstructs the
state and input. Ren \emph{et al.} train a neural network to imitate an
already-designed flatness-based feedback controller
\cite{ren2021learning}. NDF instead learns a plant-side inverse map that can be
reused inside different planning objectives, and a separate controller still
tracks the resulting reference. Do \emph{et al.} retain a known flatness
transformation and learn the geometry of its transformed constraint set
\cite{do2025ann}. NDF targets the transformation itself and handles constraints
through decoded physical quantities and audited margins. Output
identification, learned constraint sets, and NDF are therefore complementary
when more than one component of the pipeline is unavailable.

The closest structural alternative is flatness-preserving residual learning.
Yang, Welde, and Matni add a learned forward-dynamics residual to a nominal
pure-feedback system and prove that the original flat outputs are preserved
when the residual has a lower-triangular dependence structure
\cite{yang2025flatness}. Hsieh \emph{et al.} specialize this principle to
aerodynamic interaction in tight quadrotor formations and demonstrate it in
real-time hardware experiments \cite{hsieh2026formation}. These methods retain
an analytical nominal model and obtain an exact structural flatness guarantee
for the admitted residual class. They are especially attractive when the plant
can be placed in the required form and the missing physics can be represented
by the prescribed residual structure.

NDF learns a different object. It directly approximates
\(D:\xijet\mapsto(x,u)\) and embeds the frozen differentiable map in trajectory
optimization. This avoids first converting the complete plant to
pure-feedback form and does not require a lower-triangular residual, which is
useful for the coupled tail-sitter construction considered here. The broader
inverse-map interface comes with a weaker theoretical claim. NDF assumes that
an exact local inverse map already exists and proves trajectory consistency,
constraint preservation, and planning convergence only when the learned map
and its first derivative approximate that reference uniformly. It does not
claim that learning creates or globally preserves differential flatness.

\section{Learned Inverse Maps and Trajectory Optimization}
\label{sec:problem}

\subsection{Learned Inverse Map}

Consider a continuous-time control system
\begin{equation}
    \dot{x}=f(x,u), \qquad x\in\R^n,\quad u\in\R^m,
    \label{eq:dynamics}
\end{equation}
and a designer-specified candidate output \(y=h(x)\in\R^p\). Let
\begin{equation}
    \xijet=[y^\top,\dot{y}^\top,\ldots,(y^{(r)})^\top]^\top
    \in\R^q, \qquad q=p(r+1),
    \label{eq:jet}
\end{equation}
be the finite output jet required to reconstruct the system variables. If the
candidate output is flat on an operating domain \(\mathcal K\), there exists an
exact inverse map \(D^\star:\mathcal K\rightarrow\R^n\times\R^m\) satisfying
\((x,u)=D^\star(\xijet)\). Our goal is not to identify a closed-form expression
for \(D^\star\), but to approximate it from feasible data.

We use \(D\) to denote an arbitrary inverse map and \(D_\theta\) to denote its
neural approximation. Here, \(\theta\) is the collection of all trainable
network parameters, including weights and biases. The two output blocks
\(D_{\theta,x}:\R^q\rightarrow\R^n\) and
\(D_{\theta,u}:\R^q\rightarrow\R^m\) reconstruct the state and input,
respectively, according to
\begin{equation}
    (\hat{x},\hat{u})=D_\theta(\xijet)
    =\big(D_{\theta,x}(\xijet),D_{\theta,u}(\xijet)\big).
    \label{eq:decoder}
\end{equation}
Thus, \(D_\theta\) is a static inverse parameterization of a complete feasible
motion, rather than a one-step predictor of \(x(t+\Delta t)\).

Let
\(\mathcal D_N=\{(\xijet_i,x_i,u_i)\}_{i=1}^{N}\) be samples obtained from
feasible demonstration trajectories, and let \(P_{\mathcal D}\) denote their
sampling distribution on \(\mathcal K\). A generic supervised objective is
\begin{equation}
\begin{aligned}
 \mathcal{L}_{\mathrm{supervised}}(\theta)
 &=\mathbb{E}_{P_{\mathcal D}}\!\left[
   \|x-D_{\theta,x}(\xijet)\|_{W_x}^{2}\right.\\
 &\hspace{22mm}\left.{}+\|u-D_{\theta,u}(\xijet)\|_{W_u}^{2}\right].
\end{aligned}
    \label{eq:supervised}
\end{equation}
The expectation gives the population-average reconstruction error
under the sampling distribution. 

\subsection{Dynamics Consistency and Conditional Guarantees}

Let \(\xijet:[0,T]\rightarrow\mathcal K\) be an absolutely continuous jet
curve. Here and below, \(D_{\theta,x}\) denotes the assembled physical state
map, including any state components copied directly from the jet. By the chain
rule, its decoded state trajectory satisfies
\begin{equation}
 \dot{\hat{x}}(t)
 =J_{\xijet}D_{\theta,x}(\xijet(t))\dot{\xijet}(t).
 \label{eq:decoded_derivative}
\end{equation}
With \(\hat x=D_{\theta,x}(\xijet)\) and
\(\hat u=D_{\theta,u}(\xijet)\), the resulting full dynamics defect is
\begin{equation}
 \rho_\theta(\xijet,\dot{\xijet})=
 J_{\xijet}D_{\theta,x}(\xijet)\dot{\xijet}-f(\hat x,\hat u).
 \label{eq:dynamics_defect}
\end{equation}
When \(\dot{\xijet}\) and a trusted dynamics model are available, this defect
defines the generic training objective
\begin{equation}
 \mathcal{L}_{\mathrm{dyn}}(\theta)
 =\mathbb{E}_{(\xijet,\dot{\xijet})\sim P_{\mathcal D}}
 \|\rho_\theta(\xijet,\dot{\xijet})\|_{W_{\mathrm{dyn}}}^{2},
 \label{eq:generic_dyn}
\end{equation}
where \(W_{\mathrm{dyn}}\succeq0\) balances the components of the physical
residual. The derivative \(\dot{\xijet}\) contains the time derivative of the
highest-order jet component and must therefore be supplied by the trajectory
representation, an analytical data source, or numerical differentiation.

If only selected trusted equations are available, a residual
\(\mathcal R(\xijet,D_\theta(\xijet))\) certifies consistency with those
equations rather than with the full defect \(\rho_\theta\). The full-defect
bound below therefore applies when \(f\) is available and its stated uniform
approximation conditions hold.

For the guarantees below, the assembled map \(D_{\theta,x}\) is assumed to be
\(C^1\) on a neighborhood of \(\mathcal K\), so
\eqref{eq:decoded_derivative} holds almost everywhere along admissible jets.

The analysis proceeds in four steps. Assumption~\ref{ass:chart} first fixes a
regular local chart and an exact reference map \(D^\star\). Learning is not used
to establish flatness itself. Proposition~\ref{prop:defect} then shows that
uniform map-value and Jacobian errors control the output and dynamics defects
of every fixed admissible jet curve. Corollary~\ref{cor:margin} converts the
same errors into preservation of path constraints that possess a strict
margin. Section~III-C subsequently allows the optimizer to change the jet
curve. Theorem~\ref{thm:argmin} establishes convergence of planning objectives
and their approximate global minimizers, while
Proposition~\ref{prop:cover} states a finite-cover condition sufficient for the
uniform-error premise. The logical direction is therefore from a regular
chart, through \(C^1\) map approximation, to trajectory consistency and robust
feasibility, and finally to planning consistency. The finite-cover result
explains what an empirical certificate would additionally need to verify.

\begin{assumption}[regular local inverse chart]
\label{ass:chart}
The compact set \(\mathcal K\) lies in one nonsingular flatness chart. On an
open neighborhood of \(\mathcal K\), there is a single-valued inverse map
\(D^\star=(D_x^\star,D_u^\star)\in C^1\) such that, for every admissible jet
curve \(\xijet\in W^{1,\infty}([0,T],\mathcal K)\),
\begin{align}
 h(D_x^\star(\xijet(t)))&=y(t),\label{eq:leftinverse}\\
 J D_x^\star(\xijet(t))\dot{\xijet}(t)
 &=f(D_x^\star(\xijet(t)),D_u^\star(\xijet(t)))
 \label{eq:exactconsistency}
\end{align}
almost everywhere. Moreover, \(\|\dot{\xijet}\|\leq B\), and, on the relevant
compact neighborhoods, there are fixed constants \(L_x,L_u,L_h\) such that
\begin{align}
 \|f(x,u)-f(x',u')\|
 &\leq L_x\|x-x'\|+L_u\|u-u'\|,\nonumber\\
 \|h(x)-h(x')\|&\leq L_h\|x-x'\|.
 \label{eq:lipschitz_definitions}
\end{align}
\end{assumption}

Assumption~\ref{ass:chart} is deliberately local. It is the usual analytical
flatness premise restricted to the audited operating envelope. Learning does
not create this premise. Singular points and branch ambiguities must be
excluded from \(\mathcal K\) or represented by separate charts. The constants
\(L_x,L_u,L_h\) are uniform on common compact neighborhoods containing the
images of \(D^\star\) and the learned maps. Smooth activations and regularized
LayerNorm preserve local differentiability. For the tail-sitter 6-D rotation
head, the \(C^1\) premise additionally excludes zero or collinear
pre-orthogonalized columns.

\begin{proposition}[decoder-induced trajectory defect]
\label{prop:defect}
Under Assumption~\ref{ass:chart}, suppose the learned map satisfies the uniform
bounds
\begin{align}
 \|D_{\theta,x}-D_x^\star\|_{\infty,\mathcal K}&\leq\epsilon_x,&
 \|D_{\theta,u}-D_u^\star\|_{\infty,\mathcal K}&\leq\epsilon_u,\nonumber\\
 \|J D_{\theta,x}-J D_x^\star\|_{\infty,\mathcal K}
 &\leq\epsilon_J.&&\label{eq:c1errors}
\end{align}
Then every admissible jet curve obeys
\begin{align}
 \sup_t\|h(D_{\theta,x}(\xijet(t)))-y(t)\|
 &\leq L_h\epsilon_x,\label{eq:outputbound}\\
 \mathop{\mathrm{ess\,sup}}_t
 \|J D_{\theta,x}\dot{\xijet}
 -f(D_{\theta,x},D_{\theta,u})\|
 &\leq B\epsilon_J+L_x\epsilon_x+L_u\epsilon_u.
 \label{eq:bound}
\end{align}
Thus \(C^1\) convergence of the state map and uniform convergence of the input
map imply vanishing output and dynamics defects on the whole admissible
trajectory family.
\end{proposition}

\begin{IEEEproof}
By \eqref{eq:leftinverse} and the Lipschitz property of \(h\),
\begin{align}
 \|h(D_{\theta,x})-y\|
 &=\|h(D_{\theta,x})-h(D_x^\star)\|
 \leq L_h\epsilon_x.
\end{align}
For the dynamics defect, abbreviate \(x^\star=D_x^\star(\xijet)\) and
\(u^\star=D_u^\star(\xijet)\). Equation~\eqref{eq:exactconsistency} gives
\begin{equation}
 \rho_\theta=(J D_{\theta,x}-J D_x^\star)\dot{\xijet}
 +f(x^\star,u^\star)-f(\hat x,\hat u).
 \label{eq:defect_decomposition}
\end{equation}
The triangle inequality, submultiplicativity, \(\|\dot{\xijet}\|\leq B\),
and \eqref{eq:lipschitz_definitions} then yield
\begin{equation}
 \|\rho_\theta\|
 \leq B\epsilon_J+L_x\epsilon_x+L_u\epsilon_u.
\end{equation}
Taking the essential supremum over time completes the proof.
\end{IEEEproof}

\begin{corollary}[robust constraint preservation]
\label{cor:margin}
Let a decoded path constraint be \(c_j(x,u)\leq0\), where \(c_j\) is
\(L_{c_j}\)-Lipschitz. If the analytical decoding of a fixed candidate
trajectory has margin \(c_j(D^\star(\xijet(t)))\leq-\gamma_j\) for every
\(t\), then the learned decoding satisfies the same constraint whenever
\begin{equation}
 L_{c_j}(\epsilon_x+\epsilon_u)<\gamma_j.
 \label{eq:margin}
\end{equation}
For a dynamics-residual limit \(\bar r\), it is sufficient that the right-hand
side of \eqref{eq:bound} is below \(\bar r\).
\end{corollary}

\begin{IEEEproof}
For every \(t\), Lipschitz continuity and \eqref{eq:c1errors} give
\begin{align}
 c_j(D_\theta(\xijet(t)))
 &\leq c_j(D^\star(\xijet(t)))
 +L_{c_j}(\epsilon_x+\epsilon_u)\nonumber\\
 &\leq-\gamma_j+L_{c_j}(\epsilon_x+\epsilon_u)<0.
\end{align}
The residual statement follows directly from \eqref{eq:bound}.
\end{IEEEproof}

These guarantees are conditional. A small empirical value of
\(\mathcal L_{\mathrm{dyn}}\) is an average statement over sampled directions.
It does not imply the uniform value and Jacobian bounds in
\eqref{eq:c1errors}. Establishing those bounds requires additional derivative
information and coverage control. The finite-cover condition introduced below
makes this distinction explicit.

\subsection{Decoder-in-the-Loop Trajectory Optimization}
\label{sec:optimization}

\subsubsection{Spline parameterization}

Let \(0=t_0<\cdots<t_M=T\), \(h_m=t_{m+1}-t_m\), and
\(\tau_m=(t-t_m)/h_m\). On segment \(m\), the candidate output is
\begin{equation}
 y_p(t)=\sum_{j=0}^{d_s}c_{m,j}\tau_m^j,\qquad
 t\in[t_m,t_{m+1}],
 \label{eq:spline}
\end{equation}
where the free coefficients are collected in \(p\in\mathcal P\). Waypoint and
boundary equalities are absorbed into \(\mathcal P\), and adjacent segments
satisfy
\begin{equation}
 y_{p,m}^{(j)}(t_{m+1})=y_{p,m+1}^{(j)}(t_{m+1}),
 \quad j=0,\ldots,r.
 \label{eq:splinecontinuity}
\end{equation}
The degree \(d_s\) and continuity order \(r\) are chosen to provide the highest
derivative consumed by the decoder. The resulting jet is
\begin{equation}
 \xijet_p(t)=\Xi(p,t)
 =[y_p^\top,\dot y_p^\top,\ldots,(y_p^{(r)})^\top]^\top .
 \label{eq:splinejet}
\end{equation}
Auxiliary output coordinates are parameterized in the same way. A
geometry-only solution supplies the initial value \(p_0\).

\subsubsection{Planning objective and numerical audit}

For an inverse map \(D=(D_x,D_u)\), define
\begin{equation}
 x_D(p,t)=D_x(\Xi(p,t)),\qquad
 u_D(p,t)=D_u(\Xi(p,t)).
 \label{eq:decodedplan}
\end{equation}
The intended continuous-time planning problem is
\begin{align}
 \min_{p\in\mathcal P}\quad&
 \mathcal J_{\mathrm{smooth}}(p)
 \nonumber\\
 \mathrm{s.t.}\quad&
 c_j(x_D(p,t),u_D(p,t))\leq0,\quad j=1,\ldots,n_c,
 \nonumber\\
 &\mathcal R(\Xi(p,t),D(\Xi(p,t)))=0,
 \nonumber\\
 &\Xi(p,t)\in\mathcal K,\qquad t\in[0,T].
 \label{eq:constrainedplanning}
\end{align}
The current implementation evaluates these relations on a collocation grid
\(\mathcal C\) and minimizes the differentiable penalty relaxation
\begin{equation}
\mathcal{J}_{D}=\mathcal{J}_{\mathrm{smooth}}
+\rho_{\mathrm{act}}\mathcal{J}_{\mathrm{decoded}}
+\rho_{\mathrm{dyn}}\mathcal{J}_{\mathrm{dyn}}
+\rho_{\mathrm{sup}}\mathcal{J}_{\mathrm{support}}.
\label{eq:planning}
\end{equation}
With \([a]_+=\max(a,0)\), representative terms are
\begin{align}
 \mathcal J_{\mathrm{decoded}}
 &=\frac{1}{|\mathcal C|}\sum_{t_i\in\mathcal C}
   \sum_{j=1}^{n_c}[c_j(x_D,u_D)]_+^2,\nonumber\\
 \mathcal J_{\mathrm{dyn}}
 &=\frac{1}{|\mathcal C|}\sum_{t_i\in\mathcal C}
   \|\mathcal R(\Xi,D(\Xi))\|_{W_{\mathrm{dyn}}}^2,
 \label{eq:planningterms}
\end{align}

To discourage optimizer-driven extrapolation, let
\(\mathcal S_{\mathrm{tr}}\) be the training jets and define
\begin{equation}
 \ell_k=\min_{\xijet\in\mathcal S_{\mathrm{tr}}}\xi_k,\quad
 u_k=\max_{\xijet\in\mathcal S_{\mathrm{tr}}}\xi_k,\quad
 d_k=\max(u_k-\ell_k,\varepsilon).
 \label{eq:supportbounds}
\end{equation}
The normalized violation and its collocation penalty are
\begin{align}
 s_k(\xijet)
 &=\relu\!\left(\frac{\ell_k-\xi_k}{d_k}\right)
  +\relu\!\left(\frac{\xi_k-u_k}{d_k}\right),\nonumber\\
 \mathcal J_{\mathrm{support}}
 &=\frac{1}{|\mathcal C|}\sum_{t_i\in\mathcal C}\sum_k
 s_k(\Xi(p,t_i))^2.
 \label{eq:discretesupport}
\end{align}
For \(\mathcal B=\prod_k[\ell_k,u_k]\),
\begin{equation}
 \xijet\in\mathcal B\not\Rightarrow
 \operatorname{dist}(\xijet,\mathcal S_{\mathrm{tr}})
 \leq\varepsilon_{\mathrm{near}},
 \label{eq:boxlimitation}
\end{equation}
so \(\mathcal B\) is a coordinatewise guard rather than a joint-density model.

Adam minimizes \eqref{eq:planning}. On a denser audit grid \(\mathcal A\), let
\(\Xi_i=\Xi(p,t_i)\) and
\(q_i=\|\mathcal R(\Xi_i,D(\Xi_i))\|_{W_{\mathrm{dyn}}}\). We report
\begin{align}
 v_{j,\max}
 &=\max_{t_i\in\mathcal A}[c_j(x_D,u_D)]_+,
 \quad j=1,\ldots,n_c,\nonumber\\
 r_{95}&=\operatorname{Quantile}_{0.95}\{q_i:t_i\in\mathcal A\},
 \qquad
 r_{\max}=\max_{t_i\in\mathcal A}q_i,
 \label{eq:denseaudit}
\end{align}
and the support diagnostics
\begin{equation}
 r_{\mathrm{out}}=\frac{1}{|\mathcal A|}\sum_{t_i\in\mathcal A}
 \mathbf 1\{\Xi_i\notin\mathcal B\},\qquad
 s_{\max}=\max_{t_i\in\mathcal A,k}s_k(\Xi_i).
 \label{eq:supportaudit}
\end{equation}
Here \(v_{j,\max}\) preserves the units of each constraint, while \(r_{95}\)
and \(r_{\max}\) audit the selected residual \(\mathcal R\), not necessarily
the full defect \(\rho_\theta\). These finite-grid quantities measure sampled
feasibility rather than provide a solver-level guarantee.

\subsubsection{Conditional planning consistency}

Unlike the fixed-curve result in Section~III-B, the optimizer may select a
different \(p\) after replacing \(D^\star\) by \(D_\theta\). Assume
\(\mathcal P\) is compact, \(\Xi,\dot\Xi\) are continuous, and
\(\Xi(\mathcal P\times[0,T])\subset\mathcal K\). Define
\begin{equation}
 z_D(p,t)=\big(D_x(\Xi),D_u(\Xi),J D_x(\Xi)\dot\Xi\big)
 \label{eq:planningfeatures}
\end{equation}
and write \eqref{eq:planning} in the generic form
\begin{equation}
 \mathcal J_D(p)=\Phi(p)+\int_0^T
 \ell(t,p,z_D(p,t))\,dt,
 \label{eq:genericplanningobjective}
\end{equation}
where \(\Phi\) is decoder-independent and \(\ell\) is
\(L_\ell\)-Lipschitz in \(z_D\) on the reached compact set.

\begin{theorem}[objective and minimizer consistency]
\label{thm:argmin}
Let \(D_k\) be learned decoders whose errors
\((\epsilon_{x,k},\epsilon_{u,k},\epsilon_{J,k})\) in
\eqref{eq:c1errors} converge to zero, and define
\begin{equation}
 \delta_k=T L_\ell
 (\epsilon_{x,k}+\epsilon_{u,k}+B\epsilon_{J,k}).
 \label{eq:objectivegap}
\end{equation}
Then
\begin{equation}
 \sup_{p\in\mathcal P}|\mathcal J_{D_k}(p)-
 \mathcal J_{D^\star}(p)|\leq\delta_k\rightarrow0.
 \label{eq:uniformobjective}
\end{equation}
Consequently, the optimal values converge. If \(p_k\) is an
\(\eta_k\)-global minimizer,
\begin{equation}
 \mathcal J_{D_k}(p_k)
 \leq\inf_{p\in\mathcal P}\mathcal J_{D_k}(p)+\eta_k,
 \qquad \eta_k\rightarrow0,
 \label{eq:etaglob}
\end{equation}
then every accumulation point of \(\{p_k\}\) minimizes
\(\mathcal J_{D^\star}\). Under the additional quadratic-growth condition
\begin{equation}
 \mathcal J_{D^\star}(p)\geq\mathcal J_{D^\star}(p^\star)
 +\frac{\mu}{2}\|p-p^\star\|^2,
 \label{eq:growth}
\end{equation}
then
\begin{equation}
 \|p_k-p^\star\|
 \leq\sqrt{\frac{2(2\delta_k+\eta_k)}{\mu}}.
 \label{eq:minimizerbound}
\end{equation}
\end{theorem}

The proof is given in Appendix~\ref{app:proofs}. In
\eqref{eq:minimizerbound}, \(\delta_k\) measures map-replacement error and
\(\eta_k\) measures solver error. We do not claim that Adam drives \(\eta_k\)
to zero.

\begin{proposition}[finite-cover sufficient condition]
\label{prop:cover}
Let \(g:\mathcal K\rightarrow\R_{\geq0}\) be an \(L_g\)-Lipschitz audit
quantity, and let \(S_N\subset\mathcal K\) be a \(\rho_N\)-net. Every point of
\(\mathcal K\) then lies within \(\rho_N\) of some point in \(S_N\). We have
\begin{equation}
 \sup_{\xijet\in\mathcal K}g(\xijet)
 \leq \max_{\xijet_i\in S_N}g(\xijet_i)+L_g\rho_N.
 \label{eq:coverbound}
\end{equation}
Hence, if the audited value-plus-Jacobian error has a uniform Lipschitz bound,
its sampled maximum tends to zero, and \(\rho_N\rightarrow0\), then the
\(C^1\) premise of Proposition~\ref{prop:defect} and
Theorem~\ref{thm:argmin} follows.
\end{proposition}

Thus a finite audit controls the continuum through the following two terms.
\begin{equation}
 \underbrace{\max_{\xijet_i\in S_N}g(\xijet_i)}_{\text{sampled fit}}
 \;+\;
 \underbrace{L_g\rho_N}_{\text{coverage gap}}.
 \label{eq:fitcoverage}
\end{equation}
The experiments do not estimate the joint-space \(\rho_N\), \(L_g\), or
worst-case Jacobian error, so they provide empirical evidence rather than the
certificate in \eqref{eq:fitcoverage}.

\section{Experimental Evaluation}
\label{sec:experiments}

\subsection{System-Specific Implementations}
\label{sec:systems}

The following two systems instantiate the candidate output, decoder, residual,
and spline choices introduced in Section~III. They are implementation cases,
not additional assumptions in the preceding general analysis.
For \(S\in\{A,T\}\), \(\mu_S\) and \(\sigma_S\) denote training-split input
statistics, and \(\oslash\) denotes componentwise division.

The pair was selected to span complementary inverse-map regimes. The
Ackermann car has a low-dimensional, single-chart kinematic inverse, which
provides a transparent test of decoder replacement and support-aware
optimization. The tail-sitter has coupled six-DoF dynamics, continuous
rotation outputs, and hover-to-cruise chart ambiguity. It tests whether the
same interface extends to a strongly nonlinear full-envelope system rather
than reflecting properties of one vehicle class.

\begin{figure}[t]
    \centering
    \begin{minipage}[c]{0.48\columnwidth}
        \centering
        \includegraphics[width=\linewidth,trim=145 45 145 35,clip]
        {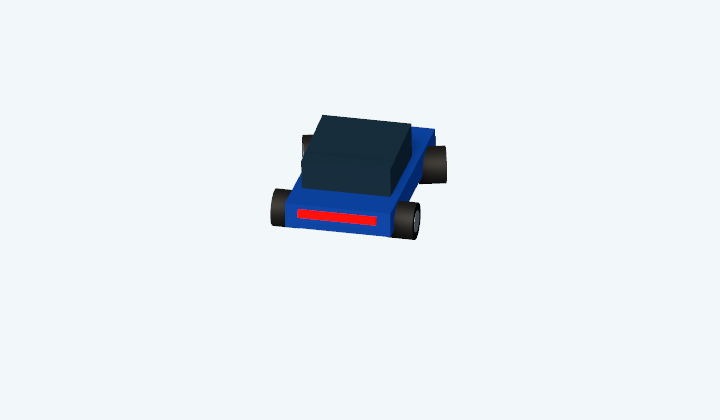}\\[-1mm]
        \footnotesize (a) Ackermann car
    \end{minipage}\hfill
    \begin{minipage}[c]{0.48\columnwidth}
        \centering
        \includegraphics[width=\linewidth,trim=145 45 145 35,clip]
        {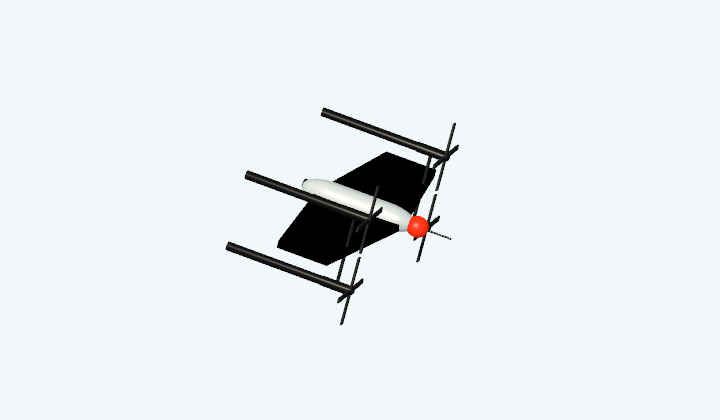}\\[-1mm]
        \footnotesize (b) Tail-sitter
    \end{minipage}
    \caption{Unity simulation objects used in the experiments. The rendered
    models provide visualization only. Training labels are generated from the
    data sources described below.}
    \label{fig:systems}
\end{figure}

Figure~\ref{fig:systems} shows the two Unity simulation objects used for
trajectory visualization. The ground vehicle is an Ackermann-steered car, and the
aircraft utilizes the Hong Hu tail-sitter model introduced in \cite{lu2024aggressive}.

\subsubsection{Ackermann Car}

Let \(y_A=\p=[p_x,p_y]^\top\). Away from zero speed, the decoder input is
\begin{equation}
    \xijet_A=[v_x,v_y,a_x,a_y]^\top,
\end{equation}
and its output is
\begin{equation}
 D_A(\xijet_A)
 =[\cos\psi,\sin\psi,\omega,a_{\parallel}]^\top.
 \label{eq:ackmap}
\end{equation}
Here \(a_{\parallel}=\dot{s}=b^\top\avec
=\vct^\top\avec/\|\vct\|_2\) is the longitudinal acceleration along the
vehicle heading, where \(s=\|\vct\|_2\) and
\(b=[\cos\psi,\sin\psi]^\top\).
The heading pair is normalized. This avoids the \(2\pi\) discontinuity and
also makes the learned forward direction explicit, eliminating the steering
sign ambiguity that occurs if visual wheel angles are inferred in a different
coordinate convention.

With \(b_\perp=[-\sin\psi,\cos\psi]^\top\), the residuals are
\begin{align}
 r_v^A &= \vct-sb,\\
 r_a^A &= \avec-a_{\parallel}b-s\omega b_\perp.
 \label{eq:ackres}
\end{align}
The network is a \(4\)-128-128-128-\(4\) multilayer perceptron (MLP) with
\(\tanh\) activations. At zero speed, heading is not determined by position
derivatives. Such samples require a separate boundary condition and are not
used to claim a globally unique chart.

For fixed waypoint-time pairs \((\bar\p_k,t_k)\), set \(d_s=5\) in
\eqref{eq:spline} and collect the free nodal derivatives as
\begin{equation}
 \vartheta_A=\{\vct_k,\avec_k\}_{k=1}^{M-1}.
 \label{eq:ackdecision}
\end{equation}
At each \(t_i\in\mathcal C_A\), let
\begin{equation}
 (b_i,\omega_i,a_{\parallel,i})=D_A(\xijet_{A,i}),\quad
 s_i=\|\vct_i\|_2,\quad
 \kappa_i=\frac{\omega_i}{\max(s_i,\varepsilon)}.
 \label{eq:ackplanquantities}
\end{equation}
Writing \(\langle z_i\rangle_A=|\mathcal C_A|^{-1}
\sum_{t_i\in\mathcal C_A}z_i\), the fixed-lemniscate implementation solves
\begin{align}
 \min_{\vartheta_A}\quad &\mathcal J_A(\vartheta_A)\nonumber\\
 \mathrm{s.t.}\quad
 &\p(t_k)=\bar\p_k,\qquad k=0,\ldots,M,\nonumber\\
 &[\p^{(q)}]_{t_k^-}^{t_k^+}=0,\quad q=0,1,2,\nonumber\\
 &\p^{(q)}(t_0)=\p^{(q)}(T)=0,\quad q=1,2,
 \label{eq:ackplanningproblem}
\end{align}
where
\begin{align}
 \mathcal J_A={}&
 \mathcal J_{A,*}
 +\lambda_{A,\mathrm{lim}}\mathcal J_{A,\mathrm{lim}}
 +\lambda_{A,\kappa}\mathcal J_{A,\kappa}\nonumber\\
 &+\lambda_{A,\mathrm{dyn}}\mathcal J_{A,\mathrm{dyn}}
 +\lambda_{A,\mathrm{dom}}\mathcal J_{A,\mathrm{dom}},
 \label{eq:ackplanningobjective}
\end{align}
with the component costs
\begin{align}
 \mathcal J_{A,*}
 &=\lambda_{A,j}\langle\|\jvec_i\|_2^2\rangle_A,\nonumber\\
 \mathcal J_{A,\mathrm{lim}}
 &=\Big\langle[s_i-\bar s_A]_+^2
 +[a_{\parallel,i}-\bar a_A]_+^2
 +[\underline a_A-a_{\parallel,i}]_+^2\Big\rangle_A,\nonumber\\
 \mathcal J_{A,\kappa}
 &=\langle m_i[|\kappa_i|-\eta_A\bar\kappa_A]_+^2\rangle_A,\nonumber\\
 \mathcal J_{A,\mathrm{dyn}}
 &=\langle\|r_{v,i}^A\|_2^2+\|r_{a,i}^A\|_2^2\rangle_A,\nonumber\\
 \mathcal J_{A,\mathrm{dom}}
 &=\langle\|\delta_{A,i}\|_2^2\rangle_A .
 \label{eq:ackplanningdefs}
\end{align}
Here
\begin{equation}
 \begin{aligned}
 m_i&=\mathbf 1\{s_i>s_\epsilon\},\\
 \delta_{A,i}&=
 \left[\left|(\xijet_{A,i}-\mu_A)\oslash\sigma_A\right|
 -\bar z_A\mathbf 1\right]_+ .
 \end{aligned}
 \label{eq:ackguarddefs}
\end{equation}
The five terms respectively penalize jerk, decoded limits, curvature,
kinematic inconsistency, and standardized extrapolation. The factor
\(\eta_A\in(0,1]\) supplies a collocation margin relative to
\(\bar\kappa_A\). Physical bounds are rechecked on the dense audit grid.

\subsubsection{Tail-sitter Aircraft}

We use the coordinated-flight analytical model of
\cite{lu2024aggressive}. Its flat output is the vehicle position
\begin{equation}
 y_T=\p\in\R^3.
 \label{eq:tailflatoutput}
\end{equation}
Away from singular configurations, the coordinated-flight constraints recover
attitude and thrust from the position derivatives. Hence, yaw is not an
independent flat output. Translation invariance gives the flat-output jet
\(\zeta_T=[\vct^\top,\avec^\top,\jvec^\top]^\top\). For a memoryless decoder,
we append the continuous auxiliary chart variable
\(b_r=[\cos\psi_r,\sin\psi_r]^\top\), giving
\begin{equation}
 \xijet_T=[
 \zeta_T^\top,b_r^\top]^\top\in\R^{11}.
 \label{eq:tailinput}
\end{equation}
The variable \(b_r\) is not part of \(y_T\). It selects the body-axis direction
when the coordinated-flight construction becomes ambiguous, such as at hover
or when its defining cross product approaches zero. It does not select the
aerodynamic angle-of-attack branch.
The decoder outputs
\begin{equation}
 D_T(\xijet_T)=[T,r_6^\top,\w^\top]^\top,
 \label{eq:tailoutput}
\end{equation}
where \(T\) is collective thrust, \(r_6\) is a continuous 6-D rotation
representation, and \(\w\) is body angular velocity. The two 3-D columns in
\(r_6\) are orthogonalized to produce \(R\in\SO(3)\)
\cite{zhou2019continuity}.

The architecture respects derivative order. A
256-256-128 Mish/LayerNorm trunk receives
\([\vct,\avec,b_r]\) and predicts thrust and attitude.
Jerk enters only the body-rate head, whose hidden widths are 128 and 64.
This prevents a higher-order derivative from unnecessarily perturbing
lower-order outputs while retaining the dependence of \(\w\) on \(\jvec\).

The original analytical construction recovers angle of attack \(\alpha\) and
thrust \(T\) by solving
\begin{equation}
 \begin{aligned}
 r_t(\alpha,T)&=T\cos\alpha-D(V,\alpha)-f_t=0,\\
 r_n(\alpha,T)&=T\sin\alpha+L(V,\alpha)-f_n=0,
 \end{aligned}
 \label{eq:alpharoot}
\end{equation}
where \(f_t\) and \(f_n\) are the required force components in the
velocity--normal plane. These equations can admit more than one root. A
sequential numerical solver therefore uses the preceding \((\alpha,T)\) as its
initial guess to stay on one continuous branch. This makes root selection
history dependent, whereas NDF receives only the current jet.

NDF does not output \(\alpha\). It predicts \((T,R,\w)\) directly, so its labels
must be single valued in the current jet. We construct them offline on a
deterministic forward-flight chart. For candidates
\(\alpha_\ell\in[-\pi/2,\pi/2]\), let \(T_\ell\) be the projected thrust and set
\begin{equation}
 \begin{aligned}
 c_\ell&=r_t(\alpha_\ell,T_\ell)^2+r_n(\alpha_\ell,T_\ell)^2\\
 &\quad+\lambda_\alpha d_{\mathbb S^1}
 (\alpha_\ell,\alpha_{\rm pref})^2+\mathcal P_T(T_\ell),\\
 w_\ell&=\frac{\exp(-c_\ell/\tau)}
 {\sum_q\exp(-c_q/\tau)},\\
 \hat\alpha&=\operatorname{atan2}\!\left(
 \sum_\ell w_\ell\sin\alpha_\ell,
 \sum_\ell w_\ell\cos\alpha_\ell\right).
 \end{aligned}
 \label{eq:alphachart}
\end{equation}
The cost combines force-balance error, a local-chart preference, and a thrust
penalty. Here \(\alpha_{\rm pref}\) blends the force-aligned low-speed chart
into the low-angle forward-flight chart, while \(\mathcal P_T\) rejects
inadmissible roots. The circular mean selects a deterministic current-jet
branch, and trajectories with an excessive attitude jump are discarded. The
selected \(\alpha\) is used only to label \((T,R,\w)\). At deployment, \(D_T\)
produces these quantities without a root solve or previous attack angle.
The auxiliary \(b_r\) selects the low-speed body-axis direction, not the
attack-angle branch.

The implementation prescribes \(b_r(t)\) at task level and does not supply
\(\dot b_r\) to the decoder. The conditional theory therefore applies on the
schedule-restricted domain where the augmented current jet defines a
single-valued inverse, while \(r_R^T\) audits the remaining kinematic defect.
Hysteresis, independently varying chart dynamics, or multiple desired branches
would require \(\dot b_r\), a chart or previous-\(\alpha\) input, or recurrent
state and lie outside Assumption~\ref{ass:chart}.

Let \(e_1\) be the body thrust axis, \(m\) the mass, \(g\) the gravity vector,
and \(F_{\mathrm{aero}}\) the aerodynamic force. The force-balance
residual is
\begin{equation}
 r_v^T=\avec-g-\frac{TRe_1+F_{\mathrm{aero}}(\vct,R)}{m}.
 \label{eq:forcebalance}
\end{equation}
An attitude-kinematic audit is
\begin{equation}
 r_R^T=\veeop(R^\top\dot{R})-\w.
 \label{eq:rotres}
\end{equation}
Here \(\veeop\) is the inverse of the hat map. It satisfies
\(\veeop([q]_\times)=q\), where \([q]_\times z=q\times z\). Since
\(R^\top\dot R\in\mathfrak{so}(3)\), \(\veeop(R^\top\dot R)\) is the
body-frame angular velocity induced by \(R(t)\). Therefore, \(r_R^T\) measures
attitude-kinematic consistency.
The main training loss is
\begin{align}
\mathcal{L}_T={}&
\lambda_T\mathcal{L}_{T,\mathrm{sup}}
+\lambda_R\mathcal{L}_{R,\mathrm{sup}}
+\lambda_\omega\mathcal{L}_{\omega,\mathrm{sup}}\nonumber\\
&+\lambda_{\mathrm{axis}}\mathcal{L}_{\mathrm{axis}}
+\lambda_{\mathrm{tail}}\operatorname{CVaR}_{0.2}(\ell_R)
+\lambda_{\mathrm{dyn}}\|r_v^T\|^2 .
\label{eq:tailloss}
\end{align}
For a mini-batch of \(B\) samples, let \(\ell_R\) denote the empirical
rotation-loss variable with sample values \(\ell_{R,b}\), so that
\(\mathcal L_{R,\mathrm{sup}}=B^{-1}\sum_{b=1}^{B}\ell_{R,b}\). We use the
upper-tail convention
\begin{equation}
 \operatorname{CVaR}_{0.2}(\ell_R)
 :=\min_{\tau\in\mathbb R}
 \left\{\tau+\frac{1}{0.2B}\sum_{b=1}^{B}
 [\ell_{R,b}-\tau]_+\right\}.
 \label{eq:cvar}
\end{equation}
Thus, \(0.2\) is the tail fraction. The term corresponds to the mean loss of
the worst \(20\%\) of attitude samples in the batch. It prevents rare,
large attitude errors from being hidden by the batch mean.
The weights are \(0.5,4,2,2,2,1\), respectively. Training examples are also
augmented by horizontal rotations. Velocity, acceleration, jerk, the auxiliary
heading pair, and target attitude rotate together. Thrust and body rates remain
invariant in the adopted convention.

For fixed position-time nodes \((\bar\p_k,t_k)\), set \(d_s=7\) and optimize
\begin{equation}
 \vartheta_T=\{\vct_k,\avec_k,\jvec_k\}_{k=1}^{M-1},
 \label{eq:taildecision}
\end{equation}
An auxiliary schedule \(b_r(t)\) is prescribed only to select the decoder chart
near degenerate segments. It is not a trajectory-optimization variable. For
\(t_i\in\mathcal C_T\), let \(s_i=\|\vct_i\|_2\) and define
\begin{equation}
 (T_i,R_i,\w_i)=D_T(\xijet_{T,i}),\quad
 \gamma_i=\frac{\operatorname{tr}(R_i^\top R_{i+1})-1}{2}.
 \label{eq:tailplanquantities}
\end{equation}
With \(\langle z_i\rangle_T=|\mathcal C_T|^{-1}
\sum_{t_i\in\mathcal C_T}z_i\), each map-specific planning run solves
\begin{align}
 \min_{\vartheta_T}\quad &\mathcal J_T(\vartheta_T)\nonumber\\
 \mathrm{s.t.}\quad
 &\p(t_k)=\bar\p_k,\nonumber\\
 &[\p^{(q)}]_{t_k^-}^{t_k^+}=0,\quad q=0,1,2,3,\nonumber\\
 &\p^{(q)}(t_0)=\p^{(q)}(T)=0,\quad q=1,2,3,
 \label{eq:tailplanningproblem}
\end{align}
where
\begin{align}
 \mathcal J_T={}&
 \mathcal J_{T,*}
 +\lambda_{T,\mathrm{lim}}\mathcal J_{T,\mathrm{lim}}
 +\lambda_{T,-}\mathcal J_{T,-}\nonumber\\
 &+\lambda_{T,F}\mathcal J_{T,F}
 +\lambda_{T,R}\mathcal J_{T,R}
 +\lambda_{T,\Delta}\mathcal J_{T,\Delta}\nonumber\\
 &+\lambda_{T,\mathrm{dom}}\mathcal J_{T,\mathrm{dom}}
 +\lambda_{T,\mathrm{sup}}\mathcal J_{\mathrm{support}},
 \label{eq:tailplanningobjective}
\end{align}
and the component costs are
\begin{align}
 \mathcal J_{T,*}
 &=\lambda_{T,s}\langle\|\p_i^{(4)}\|_2^2\rangle_T,\nonumber\\
 \mathcal J_{T,\mathrm{lim}}
 &=\Big\langle[s_i-\bar s_T]_+^2+[T_i-\bar T]_+^2
 +[\|\w_i\|_2-\bar\omega]_+^2\Big\rangle_T,\nonumber\\
 \mathcal J_{T,-}
 &=\langle[\underline T-T_i]_+^2\rangle_T,\nonumber\\
 \mathcal J_{T,F}
 &=\langle\|r_{v,i}^T\|_2^2\rangle_T,\qquad
 \mathcal J_{T,R}=\langle\|r_{R,i}^T\|_2^2\rangle_T,\nonumber\\
 \mathcal J_{T,\Delta}
 &=\langle[\cos\bar\theta-\gamma_i]_+^2\rangle_T,\nonumber\\
 \mathcal J_{T,\mathrm{dom}}
 &=\langle\|\delta_{T,i}\|_2^2\rangle_T .
 \label{eq:tailplanningterms}
\end{align}
Here
\begin{equation}
 \delta_{T,i}=
 \left[\left|(\xijet_{T,i}-\mu_T)\oslash\sigma_T\right|
 -\bar z_T\mathbf 1\right]_+ .
 \label{eq:taildomainguard}
\end{equation}
These terms respectively encode snap, decoded limits, minimum thrust,
force-balance consistency, attitude-kinematic consistency, attitude
continuity, standardized extrapolation, and the empirical training box in
\eqref{eq:discretesupport}. All limits and residuals are recomputed on the
denser audit grid.

\subsubsection{Data Sources and Evaluation Protocol}

\paragraph{Ackermann generator}
Each trajectory uses four to seven forward-progressing planar nodes with
longitudinal increments drawn from \(\mathcal U(3,8)\) and lateral increments
from \(\mathcal U(-2,2)\). Natural cubic splines interpolate the nodes, and
\((\p,\vct,\avec)\) is sampled at 200 times. Frames must satisfy
\(s>0.5\), \(|\kappa|\leq\tan(35^\circ)/2.5\), and
\(-5\leq a_\parallel\leq3\). A trajectory is retained when at least 80\% of
its frames are valid. Analytical labels use
\(\psi=\operatorname{atan2}(v_y,v_x)\),
\(\omega=(v_xa_y-v_ya_x)/\|\vct\|_2^2\), and
\(a_\parallel=\vct^\top\avec/\|\vct\|_2\).

\paragraph{Tail-sitter generator}
Independently of the ground-vehicle splines, tail-sitter data combine two
mission families. Randomized phase profiles produce complete
takeoff--transition--turn--landing motions with cruise speed in \([7.5,20]\),
turn radius in \([40,70]\), and vertical displacement in \([8,45]\). A second
family fits quintic three-dimensional B-splines through transition and cruise
waypoints, then increases duration until speed and acceleration meet the
generator limits. Both families provide \((\vct,\avec,\jvec)\) along each
mission. Complete missions are augmented by time scaling
\(c\in[0.82,1.22]\), horizontal rotation, and optional lateral reflection,
with derivative orders scaled by \(c^{-1}\), \(c^{-2}\), and \(c^{-3}\).
The deterministic chart in \eqref{eq:alphachart} supplies synchronized
\((T,R,\w)\) labels, and missions violating thrust, body-rate, or
adjacent-attitude limits are rejected. The reproduction package records the
exact phase profiles and generator thresholds.

Both primary datasets are synthetic analytical-teacher data. Unity is used
only to render planned motions. Section~\ref{sec:mismatch} separately tests
noisy labels from a hidden perturbed teacher.

Table~\ref{tab:data} summarizes both datasets. Complete trajectories, rather
than frames, are assigned to a common training, validation, and test protocol,
preventing adjacent samples from one mission from leaking across splits.
Independent training repeats use the same trajectory split and protocol. The
tables report their mean and sample standard deviation.
For samplewise or trajectorywise errors, p95 denotes the empirical 95th
percentile. Thus, 95\% of the evaluated values do not exceed it. This statistic
captures difficult transition cases without being dominated by one maximum.

\begin{table}[t]
\centering
\caption{Datasets and decoder interfaces.}
\label{tab:data}
\footnotesize
\setlength{\tabcolsep}{3pt}
\begin{tabular}{@{}lcc@{}}
\toprule
 & Ackermann & Tail-sitter\\
\midrule
Flat output & position & position\\
Jet & \(\vct,\avec\) & \(\vct,\avec,\jvec\)\\
Auxiliary chart & -- & \(b_r=[\cos\psi_r,\sin\psi_r]\)\\
Decoder output & heading, \(\omega,a_\parallel\) & \(T,R,\w\)\\
Trajectories & 2000 & 720\\
Frames & 392,553 & 340,800\\
Split (traj.) & 1400/300/300 & 504/108/108\\
Test frames & 58,741 & 48,920\\
Unseen suite & 140 traj., 67,200 frames & 140 traj., 67,200 frames\\
\bottomrule
\end{tabular}
\end{table}

\subsection{Held-Out Mapping and Reproducibility}

Table~\ref{tab:repeatedruns} reports mapping accuracy. The Ackermann map is
consistently accurate on both held-out random splines and geometrically new
motions. The tail-sitter map also preserves comparable accuracy on its unseen
suite, with small variability across independent training repeats. These
results show stable interpolation and task-family transfer within the sampled
operating domain. They do not establish accuracy outside that domain.

The system-specific contents of the confirmation suites are as follows.
\begin{itemize}
    \item The Ackermann suite contains lane change, double lane change,
    S-bend, slalom, asymmetric chicane, curvature sweep, and offset return.
    All trajectories meet the heading and kinematic-residual criteria and
    remain well covered by the training distribution.
    \item The tail-sitter suite contains figure eight, descending turn,
    climbing full circle, stop--yaw--resume, helical slalom, double-turn climb,
    and accelerating \(270^\circ\) turn. Training retains generic transition
    and turn primitives, but none of these families. All trajectories meet the
    attitude and force-balance criteria. Coordinatewise box crossings are
    reported separately as coverage warnings.
\end{itemize}
Both suites contain 20 complete trajectories from each of seven families.
Their definitions and exact trajectories are excluded from training and model
selection. Thus, \textit{unseen} denotes task-family and trajectory holdout,
not that every local jet lies outside training support.

\begin{table}[t]
\centering
\caption{Fixed-split mapping accuracy.}
\label{tab:repeatedruns}
\footnotesize
\begin{tabular}{@{}llc@{}}
\toprule
System & Metric & Result\\
\midrule
\multirow{3}{*}{Ackermann}
& Head. p95 (deg) & \(0.0603\pm0.0049\)\\
& Rate RMSE (rad/s) & \(0.00204\pm0.00013\)\\
& Dyn. p95 (mixed SI) & \(0.00782\pm0.00040\)\\
\midrule
\multirow{3}{*}{Ack., unseen}
& Head. p95 (deg) & \(0.0589\pm0.0218\)\\
& Rate RMSE (rad/s) & \(0.00740\pm0.00011\)\\
& Dyn. p95 (mixed SI) & \(0.01328\pm0.00095\)\\
\midrule
\multirow{3}{*}{Tail, held-out}
& Att. p95 (deg) & \(2.606\pm0.028\)\\
& Rate p95 (rad/s) & \(0.03059\pm0.00026\)\\
& Force p95 (m/s\(^2\)) & \(0.2207\pm0.0066\)\\
\midrule
\multirow{3}{*}{Tail, unseen}
& Att. p95 (deg) & \(1.979\pm0.068\)\\
& Rate p95 (rad/s) & \(0.0679\pm0.0042\)\\
& Force p95 (m/s\(^2\)) & \(0.209\pm0.011\)\\
\bottomrule
\end{tabular}
\par\vspace{0.5mm}
\parbox{0.94\columnwidth}{\footnotesize\textit{Note.} Ackermann Dyn. is the
norm of the velocity and acceleration residual blocks in \eqref{eq:ackres}.
The blocks retain their native SI units.}
\end{table}

\subsection{Ablations}

Table~\ref{tab:ablation} first isolates five tail-sitter design choices under
the same split and training protocol. Let the full configuration be
\begin{equation}
 \mathfrak A_{\rm full}
 =\{\mathcal L_T,D_{\rm str},\mathcal G_{\rm yaw},p_{\rm hard}\},
 \label{eq:ablationfull}
\end{equation}
where \(\mathcal L_T\) is \eqref{eq:tailloss}, \(D_{\rm str}\) is the
derivative-ordered decoder, and \(\mathcal G_{\rm yaw}\) applies a common
horizontal rotation to each input--label pair. With standardized input
\(z=(\xijet_T-\mu_T)\oslash\sigma_T\), hard sampling uses
\begin{equation}
 p_{\rm hard}(i)
 \propto 1+\lambda_h
 \min\!\left\{\frac{\|z_i\|_\infty}{q_{0.99}},1\right\}^{2},
 \label{eq:hardsampling}
\end{equation}
where \(q_{0.99}\) is the training-set 99th percentile of
\(\|z\|_\infty\). The reported variants are defined by
\begin{align}
 \mathfrak A_{-\rm dyn}:&\quad
 \mathcal L_T\leftarrow\mathcal L_T-\lambda_{\rm dyn}\|r_v^T\|_2^2,
 \nonumber\\
 \mathfrak A_{\rm mono}:&\quad
 D_{\rm str}\leftarrow D_{\rm mono}:\xijet_T\mapsto(T,r_6,\w),
 \nonumber\\
 \mathfrak A_{-\rm tail}:&\quad
 \mathcal L_T\leftarrow\mathcal L_T-\lambda_{\rm axis}
 \mathcal L_{\rm axis}-\lambda_{\rm tail}
 \operatorname{CVaR}_{0.2}(\ell_R),
 \nonumber\\
 \mathfrak A_{-\rm yaw}:&\quad \mathcal G_{\rm yaw}\leftarrow I,
 \qquad
 \mathfrak A_{-\rm hard}:\quad p_{\rm hard}(i)\leftarrow N^{-1}.
 \label{eq:ablationvariants}
\end{align}
Thus, each row removes one regularizer, structural restriction, augmentation,
or sampling rule while leaving the remaining configuration unchanged.
For the Ackermann map, the applicable controlled removal is
\begin{equation}
 \mathfrak A^A_{-\rm dyn}:\quad
 \mathcal L_A\leftarrow\mathcal L_A-\lambda_{A,\rm dyn}
 \big(\|r_v^A\|_2^2+\|r_a^A\|_2^2\big).
 \label{eq:ackablation}
\end{equation}

\begin{table*}[t]
\centering
\caption{Decoder ablation studies.}
\label{tab:ablation}
\footnotesize
\textit{(a) Tail-sitter}\par\vspace{0.5mm}
\begin{tabular}{@{}lrrrrrr@{}}
\toprule
Variant &
\multicolumn{1}{c}{H-att. (deg)} &
\multicolumn{1}{c}{H-res. (m/s\(^2\))} &
\multicolumn{1}{c}{U-att. (deg)} &
\multicolumn{1}{c}{U-res. (m/s\(^2\))} &
\multicolumn{1}{c}{U-rate (rad/s)} &
\multicolumn{1}{c}{Acc. (\%)}\\
\midrule
Full & 2.620 & 0.228 & 1.979 & 0.206 & 0.072 & 100.0\\
No dynamics loss & 1.805 & 0.612 & 2.251 & 0.334 & 0.077 & 100.0\\
Monolithic decoder & 2.699 & 0.241 & 2.420 & 0.236 & 0.131 & 100.0\\
No axis/CVaR losses & 3.094 & 0.193 & 2.089 & 0.208 & 0.072 & 100.0\\
No yaw symmetry & 2.406 & 0.171 & 9.289 & 1.004 & 0.189 & 44.3\\
No hard sampling & 2.636 & 0.222 & 1.952 & 0.232 & 0.065 & 100.0\\
\bottomrule
\end{tabular}

\vspace{1.5mm}
\textit{(b) Ackermann}\par\vspace{0.5mm}
\begin{tabular}{@{}lrrrr@{}}
\toprule
Variant & Head. p95 (deg) & Rate RMSE (rad/s) &
Acc. RMSE (m/s\(^2\)) & Dyn. p95 (mixed SI)\\
\midrule
Full & 0.05935 & 0.001948 & 0.002366 & 0.007771\\
No dynamics loss & 0.06256 & 0.002079 & 0.002393 & 0.008588\\
\bottomrule
\end{tabular}
\par\vspace{0.7mm}
\parbox{0.94\textwidth}{\footnotesize\textit{Note.} H/U =
held-out/unseen. Att./res./rate/Acc. denote attitude p95, force-residual p95,
body-rate p95, and trajectory acceptance. Ackermann Dyn. is defined as in
Table~\ref{tab:repeatedruns}.}
\end{table*}

The full tail-sitter configuration is not the numerical winner in every
isolated column. Its advantage is a balanced result across label accuracy,
force consistency, rate prediction, and unseen-task acceptance. Removing the
dynamics loss reduces held-out attitude p95, but raises held-out and unseen
force-residual p95 by \(2.68\times\) and \(1.62\times\), respectively. Thus,
teacher-label fit alone does not ensure force balance. Replacing the structured
decoder with a monolithic MLP increases unseen attitude and force-residual p95
and raises rate p95 from \(0.072\) to \(0.131\). The derivative-ordered
architecture therefore provides its clearest gain in the higher-order output.

The thrust-axis and CVaR terms primarily suppress the attitude tail. Their
removal increases held-out attitude p95 from \(2.620\) to \(3.094\), while the
force residual changes little. Horizontal-rotation augmentation has the
largest generalization effect. Without it, apparently favorable held-out
numbers coexist with much larger unseen errors and only \(44.3\%\) acceptance.
Hard sampling gives a smaller tradeoff. Removing it slightly improves unseen
attitude and rate errors in this run but worsens unseen force-residual p95.
Accordingly, hard sampling is treated as a boundary-coverage aid rather than
an individually necessary component.

The Ackermann ablation shows the same direction but a smaller effect. Keeping
the dynamics term improves heading, rate, longitudinal-acceleration, and
combined-residual metrics. The changes are modest because the Ackermann
inverse is simple and single valued, so supervised labels already impose most
of the kinematic structure. This comparison explains why the richer
tail-sitter map benefits more strongly from explicit consistency training.

\subsection{Optimization from Unseen Sparse Tasks}
\label{sec:unseenopt}

The two benchmarks have the following system-specific requirements.
\begin{itemize}
    \item Ackermann tasks use only position--time nodes. Their feasibility
    protocol is frozen on a disjoint development subset.
    \item Tail-sitter tasks add a prescribed auxiliary heading schedule near
    degenerate segments and use the quality rules in the Appendix.
\end{itemize}
Otherwise, both benchmarks use 35 frozen missions, five from each family in
Section~IV-B, with 13 position--time nodes per task. Dense teacher outputs are
not used for targets or initialization, and both maps share a geometry-only
warm start. The analytical and learned maps are inserted into otherwise
identical planners. If \(\mathfrak q\) collects their common settings, the runs
are
\begin{equation}
 \vartheta_\star\in\arg\min_\vartheta\mathcal J(\vartheta;D^\star,\mathfrak q),
 \quad
 \vartheta_\theta\in\arg\min_\vartheta\mathcal J(\vartheta;D_\theta,\mathfrak q).
 \label{eq:controlledreplacement}
\end{equation}
This pairing is only an experimental control. It isolates decoder replacement
without assigning a separate methodological role to it.

\begin{figure*}[t]
    \centering
    \includegraphics[width=0.75\textwidth,height=0.60\textheight,
    keepaspectratio]{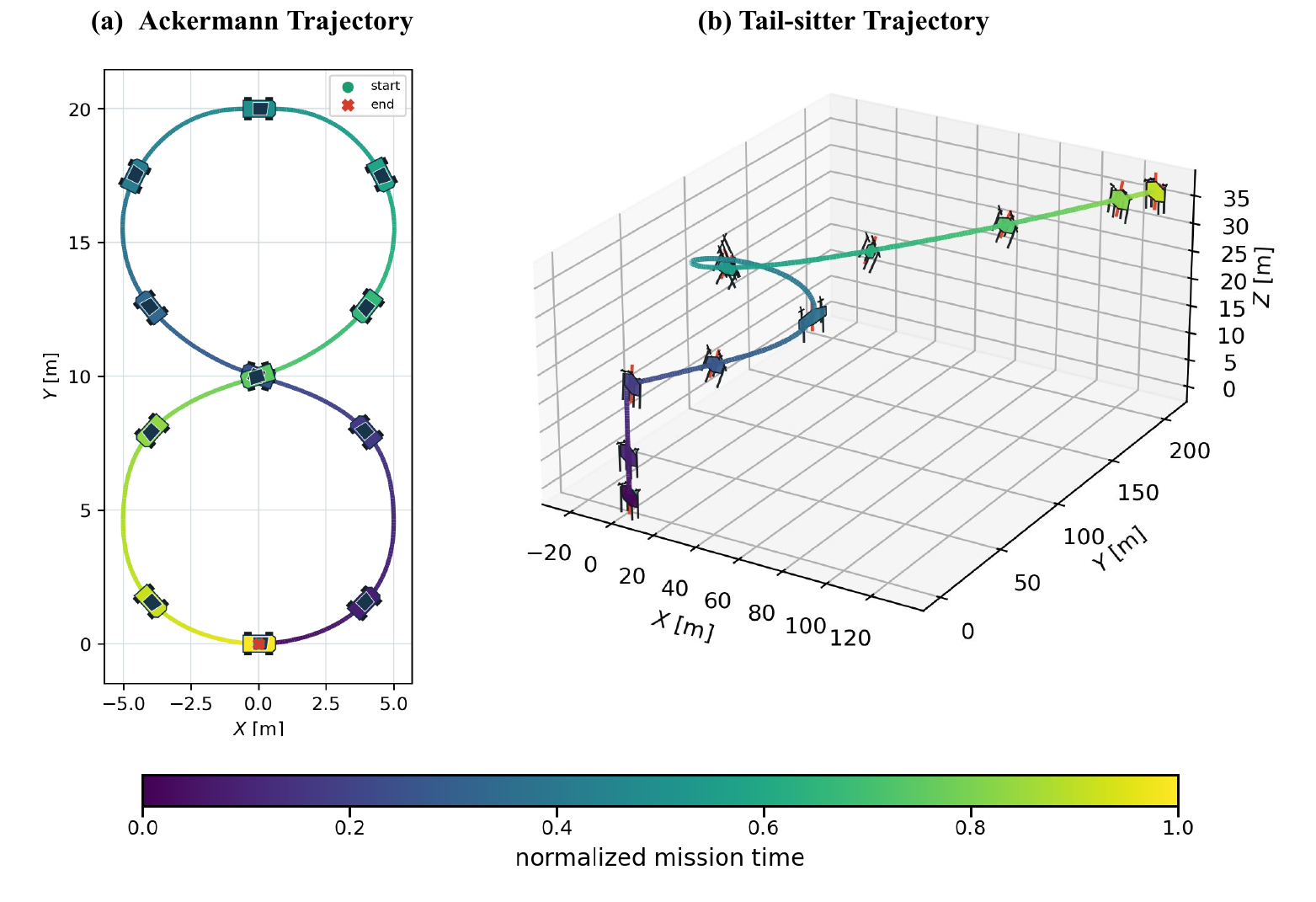}
    \caption{NDF trajectory geometry with metric coordinate axes and vehicle
    models sampled uniformly in mission time. (a) Ackermann figure-eight in
    top view. (b) Three-dimensional tail-sitter full-envelope mission.
    The model glyphs are enlarged for legibility and use the recorded
    attitudes. Color denotes normalized mission time. Only the learned path is
    shown because analytical and learned spatial paths are visually similar.
    Their decoded quantities are compared numerically in
    Table~\ref{tab:optimization}.}
    \label{fig:motion}
\end{figure*}

\begin{figure*}[t]
    \centering
    \includegraphics[width=0.85\textwidth,height=0.70\textheight,
    keepaspectratio]{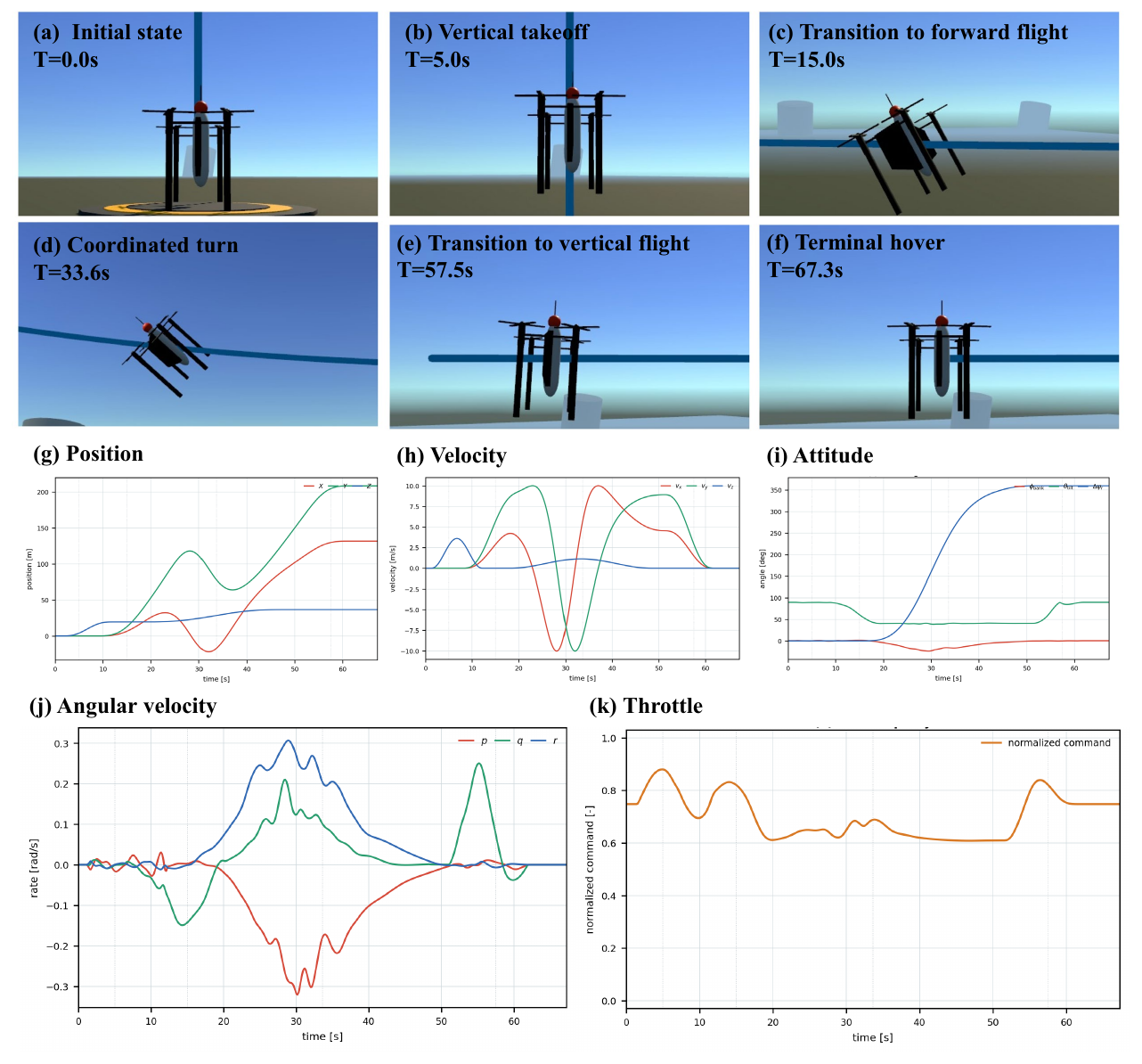}
    \caption{Unity visualization and time histories for the representative
    three-dimensional tail-sitter mission. Panels (a)--(f) show the initial
    state, vertical takeoff, transition, coordinated climbing turn,
    re-transition, and terminal hover. Panels (g)--(j) combine all three axes
    of position and velocity, task-aligned attitude coordinates, and body
    angular velocity. Dotted vertical lines correspond to the six displayed
    frames. In panel (i), \(\theta_{\rm tilt}\) is the elevation of the body
    thrust axis, \(\Delta\psi_r=\psi_r-\psi_r(0)\) is the unwrapped yaw
    reference relative to its initial value, and \(\phi_{\rm bank}\) is the
    rotation of the body span axis about a path frame constructed from
    \(\psi_r\) and the flight-path angle. Panel (k) is a scalar throttle
    proxy obtained by linearly normalizing the exported 1--40N total-thrust
    command to \([0,1]\), rather than a measured ESC duty cycle. The complete
    44.9s video is provided as a multimedia artifact.}
    \label{fig:storyboard}
\end{figure*}

Table~\ref{tab:optimization} places both unseen-task benchmarks in one table.
For both systems, replacing the analytical inverse map by NDF preserves task
success and audited physical feasibility. The learned solutions also retain
small dynamics defects, including at the across-task tail. The support results
instantiate \eqref{eq:supportaudit}. A mission is fully in support when every
dense-grid jet lies in the coordinatewise training box \(\mathcal B\), or
equivalently when \(r_{\rm out}=0\). The audit-point outside rate is the mean of
\(100r_{\rm out}\) over 35 missions. Thus, physical feasibility is distinct
from interpolation coverage, and a feasible trajectory may have a few jets
outside the conservative box. The unoptimized neural map is slower than the
analytical map, so this experiment establishes replacement reliability rather
than computational advantage over an available closed form. Constraint
margins, feasible-first selection, dense audit, and restoration remain
necessary even when the decoder residual is small.
Task p95 is taken across the 35 missions.

\begin{table}[t]
\centering
\caption{Optimization on unseen tasks.}
\label{tab:optimization}
\footnotesize
\setlength{\tabcolsep}{3pt}
\begin{tabular}{@{}llrr@{}}
\toprule
System & Metric & Analyt. & NDF\\
\midrule
\multirow{8}{*}{Ackermann}
& Success (\%) & 100 & 100\\
& Physical viol. (\%) & 0 & 0\\
& Fully in support (\%) & 97.1 & 97.1\\
& Audit points outside (\%) & 0.013 & 0.015\\
& Dyn., median (mixed SI) & \(2.63{\times}10^{-6}\) & 0.0107\\
& Dyn., task p95 (mixed SI) & \(2.91{\times}10^{-6}\) & 0.0424\\
& Time, median (s) & 17.69 & 25.47\\
& Time, task p95 (s) & 31.57 & 39.66\\
\midrule
\multirow{9}{*}{Tail-sitter}
& Success (\%) & 100 & 100\\
& Physical viol. (\%) & 0 & 0\\
& Fully in support (\%) & 51.4 & 74.3\\
& Audit points outside (\%) & 0.917 & 0.125\\
& Dyn., median (m/s\(^2\)) & 0.232 & 0.273\\
& Dyn., task p95 (m/s\(^2\)) & 0.477 & 0.465\\
& Kin., median (rad/s) & 0.097 & 0.106\\
& Time, median (s) & 14.54 & 19.14\\
& Time, task p95 (s) & 15.26 & 19.77\\
\bottomrule
\end{tabular}
\end{table}

\subsection{Receding-Horizon MPC Comparison}
\label{sec:mpccomparison}

This comparison evaluates computation--accuracy tradeoffs rather than a new
MPC formulation. We use standard constrained receding-horizon baselines
following \cite{mayne2000constrained,salzmann2023realtime}. Training time is
excluded from all online measurements.

For Ackermann motion, both methods use the same kinematic bicycle plant,
waypoints, boundary states, and speed, acceleration, and steering limits. The
baseline uses a 25-step horizon at \(0.1\) s, a fixed 40-iteration update
budget, and applies the first optimized input. NDF instead solves one
piecewise-quintic mission problem with the same nodes and physical limits.
Reported wall time includes all optimization and map evaluation after the
mission is received. For MPC it is the sum of every horizon solve. It therefore
measures mission-level solver workload rather than worst-case single-update
latency.

For the tail-sitter, both methods use the six-DoF model of
\cite{lu2024aggressive}, the same predeclared full-circle-climb reference,
geometric stabilizer, constraints, disturbances, and five fixed initial
perturbations. The NDF reference is generated once before timing and then
tracked by the stabilizer. The NMPC baseline uses warm-started direct shooting
to optimize bounded control corrections around that same stabilizer at the
measured 0.1404-s sample period. Table~\ref{tab:mpc} reports mean \(\pm\)
sample standard deviation over the five trials. Exact solver settings and the
executable benchmark are provided in the reproduction package rather than
repeated here.

\begin{table}[t]
\centering
\caption{Computational tradeoffs with receding-horizon baselines.}
\label{tab:mpc}
\footnotesize
\textit{(a) Ackermann mission-level planning}\par\vspace{0.5mm}
\begin{tabular}{@{}lrr@{}}
\toprule
Metric & NDF & MPC\\
\midrule
Solver time (s) & 1.168 & 187.734\\
Rel. workload & \(1.0\times\) & \(160.7\times\)\\
\(\int\dot\kappa^2dt\) & 0.61 & 5.38\\
Completion & yes & yes\\
\bottomrule
\end{tabular}

\vspace{1.5mm}
\textit{(b) Tail-sitter six-DoF closed-loop control}\par\vspace{0.5mm}
\begin{tabular}{@{}lrr@{}}
\toprule
Metric & NDF & NMPC\\
\midrule
Success (\%) & 100 & 100\\
Pos. p95 (m) & \(1.952\pm0.059\) & \(0.476\pm0.027\)\\
Att. p95 (deg) & \(15.822\pm1.441\) & \(5.389\pm1.176\)\\
Update p95 (ms) & \(0.123\pm0.005\) & \(21.825\pm0.449\)\\
Online total (s) & \(0.050\pm0.0005\) & \(9.790\pm0.081\)\\
Misses (\%) & 0 & 0\\
\bottomrule
\end{tabular}
\end{table}

On the Ackermann mission, both methods complete the task, while NDF requires
\(160.7\times\) less mission-level solver time and reduces curvature-rate
energy by 88.7\%. On the tail-sitter trials, NMPC uses online forward-model
optimization to reduce position and attitude errors. The NDF-reference tracker
requires \(196.7\times\) less total online computation. Both remain below the
control deadline. These results demonstrate a workload--accuracy tradeoff for
the evaluated implementations, not universal dominance over optimized MPC.

\subsection{Noisy Logs and Model Mismatch}
\label{sec:mismatch}

The analytical teacher in the preceding experiments is the ideal case under the
assumed tail-sitter model, not measured truth. We therefore construct a hidden
perturbed plant and generate 720 synthetic trajectories. Training labels
contain trajectory bias, correlated AR(1) noise (\(\rho=0.90\)--0.92), and
0.2\% bounded outliers. NDF is trained only by supervised loss in this study.
It receives neither the hidden plant parameters nor the true force residual.

\emph{Baselines and terminology.} Input standardization,
\(\tilde\xijet=(\xijet-\mu)/\sigma\), is only numerical preprocessing for NDF
and is unrelated to the \emph{nominal} analytical baseline. The nominal map
uses fixed design parameters and no deployment data. The calibrated analytical
map retains the same functional form but fits four constants from the training
logs. These constants are thrust gain, lift scale, drag scale, and drag offset.
NDF instead learns
the complete jet-to-output map without imposing this constant-parameter form.
A true-model analytical map is included only as the ideal case.

\emph{Parametric mismatch as a negative control.} When the hidden discrepancy
is predominantly low dimensional and remains inside the calibrated model
class, the calibrated analytical map is the best deployable method. This result
establishes that inexpensive parameter identification should be preferred when
the analytical structure is adequate. NDF is not introduced as a substitute
for a calibration problem that is already well posed.

\emph{Structural mismatch.} The stress test adds a speed-dependent
command-to-thrust effectiveness law, whereas calibration is restricted to one
constant gain. On 140 clean unseen trajectories, Table~\ref{tab:mismatch}
shows that noisy-log NDF reduces thrust-error p95 by 56.0\% and true dynamics-residual p95 from by 34.9\% relative to the calibrated analytical map. Relative to the nominal
map, the reductions are 92.8\% and 86.0\%, respectively. These gains arise
without exposing NDF to the hidden speed-dependent law, demonstrating the
intended advantage when the mismatch lies outside the chosen analytical model
class. The calibrated map remains more accurate in attitude and body rate,
where its geometric structure supplies a stronger prior. The result therefore
supports nonparametric force-channel adaptation, rather than uniform NDF
superiority. NDF entries are mean \(\pm\) sample standard deviation over three
runs.

\begin{table}[t]
\centering
\caption{Speed-dependent structural mismatch. Bold marks the best deployable
method in each channel.}
\label{tab:mismatch}
\footnotesize
\resizebox{\columnwidth}{!}{%
\begin{tabular}{@{}lrrrr@{}}
\toprule
Method & Att. (deg) & Thrust (N) & Rate (rad/s) & Dyn. (m/s\(^2\))\\
\midrule
Nominal analytical & 5.693 & 14.626 & 0.0242 & 3.638\\
Calibrated analytical & \textbf{1.874} & 2.406 &
\textbf{0.0114} & 0.785\\
NDF, noisy logs & \(3.123\pm0.248\) &
\(\mathbf{1.058\pm0.021}\) & \(0.114\pm0.003\) &
\(\mathbf{0.511\pm0.043}\)\\
True model & \(\approx0\) & 0 & 0 & 0.167\\
\bottomrule
\end{tabular}%
}
\end{table}

\section{Conclusion}
\label{sec:conclusion}

This paper presents a structured neural inverse-flatness map that is trained
from feasible motion data and embedded directly in flat-output trajectory
optimization. Conditional analysis connects value and Jacobian approximation
to dynamics consistency, constraint preservation, and convergence of the
planning problem. Across Ackermann and tail-sitter experiments, the learned map
is stable on held-out data, succeeds on all 70 unseen planning tasks, adapts to
a tested structural actuator mismatch, and reduces solver workload relative to
the evaluated MPC implementations. The six-DoF tail-sitter study also makes
the tradeoff explicit. NMPC tracks more accurately, while the NDF-reference
pipeline removes repeated horizon optimization.

Future work will automate selection of physical outputs and derivative orders,
develop multi-chart or history-conditioned inverse maps, and construct
finite-sample \(C^1\) and support certificates for robust planning.

\appendices

\section{Proofs of the Planning-Convergence Results}
\label{app:proofs}

\begin{IEEEproof}[Proof of Theorem~\ref{thm:argmin}]
For every \((p,t)\), the definition \eqref{eq:planningfeatures}, the uniform
decoder bounds, and \(\|\dot\Xi\|\leq B\) give
\begin{equation}
 \|z_{D_k}(p,t)-z_{D^\star}(p,t)\|
 \leq\epsilon_{x,k}+\epsilon_{u,k}+B\epsilon_{J,k}.
 \label{eq:featureboundappendix}
\end{equation}
The Lipschitz property of \(\ell\), followed by integration on \([0,T]\),
yields \eqref{eq:uniformobjective}. Taking infima on both sides shows
\begin{equation}
 \left|\inf_{p\in\mathcal P}\mathcal J_{D_k}(p)
 -\inf_{p\in\mathcal P}\mathcal J_{D^\star}(p)\right|\leq\delta_k,
 \label{eq:valueconvergenceappendix}
\end{equation}
so the optimal values converge.

Let \(p^\star\) be any minimizer of the analytical problem. Approximate
optimality and \eqref{eq:uniformobjective} imply
\begin{align}
 \mathcal J_{D^\star}(p_k)
 &\leq \mathcal J_{D_k}(p_k)+\delta_k\nonumber\\
 &\leq \inf_p\mathcal J_{D_k}(p)+\eta_k+\delta_k\nonumber\\
 &\leq \mathcal J_{D_k}(p^\star)+\eta_k+\delta_k\nonumber\\
 &\leq \mathcal J_{D^\star}(p^\star)+2\delta_k+\eta_k.
 \label{eq:argminchain}
\end{align}
Compactness of \(\mathcal P\) supplies a convergent subsequence. Passing to
its limit in \eqref{eq:argminchain}, using continuity of
\(\mathcal J_{D^\star}\), proves that the accumulation point is an analytical
global minimizer. Finally, combining \eqref{eq:argminchain} with the quadratic
growth condition \eqref{eq:growth} gives
\(\frac{\mu}{2}\|p_k-p^\star\|^2\leq2\delta_k+\eta_k\), which is
\eqref{eq:minimizerbound}.
\end{IEEEproof}

\begin{IEEEproof}[Proof of Proposition~\ref{prop:cover}]
For any \(\xijet\in\mathcal K\), choose \(\xijet_i\in S_N\) with
\(\|\xijet-\xijet_i\|\leq\rho_N\). Lipschitz continuity gives
\begin{equation}
 g(\xijet)\leq g(\xijet_i)+L_g\|\xijet-\xijet_i\|
 \leq\max_{\xijet_j\in S_N}g(\xijet_j)+L_g\rho_N.
\end{equation}
Taking the supremum over \(\mathcal K\) proves \eqref{eq:coverbound}. Apply
this inequality to the sum of the two value errors and the state-map Jacobian
error. If its sampled maximum and \(\rho_N\) both vanish while \(L_g\) remains
bounded, then all three errors in \eqref{eq:c1errors} vanish, and the preceding
results apply.
\end{IEEEproof}

\section{Detailed Reproduction Settings}

The Ackermann and tail-sitter datasets contain 2000/720 complete trajectories
and 392,553/340,800 frames, respectively. Each unseen confirmation suite adds
140 trajectories and 67,200 frames. Both systems use trajectory-grouped
70/15/15\% train/validation/test splits. Standardization and support bounds use
training groups only.

Every unseen mission has 13 sparse nodes and 12 segments. Ackermann uses a
100-iteration geometry warm start, 800 map-specific iterations, and a
200-iteration feasibility phase. Tail-sitter uses 600 and 800 iterations,
respectively. The analytical and learned runs share initialization, loss
weights, collocation, and audit grids, with execution order alternated by task.
Success requires completion, zero audited physical-limit violation, and the
predeclared residual and attitude-continuity thresholds stated with the result
tables. Exact weights, learning rates, dense-grid sizes, and executable
commands are archived in the accompanying reproduction package. Omitting them
here avoids duplicating machine-readable configuration.

\section{Synthetic Log-Noise Configuration}

The hidden plant changes mass, aerodynamic scales, drag offset, thrust gain,
and angle-dependent lift/drag terms. The structural stress test additionally
uses a speed-dependent thrust-effectiveness slope of 0.35. Inputs and labels
receive trajectory bias, AR(1) noise with correlations 0.90--0.92, and 0.2\%
bounded outliers. Noise scales and all hidden parameters are recorded in the
reproduction package. These synthetic settings isolate an adaptation mechanism
and are not identified Hong Hu uncertainty.

\bibliographystyle{IEEEtran}
\bibliography{references}

\begin{thebibliography}{10}
\providecommand{\url}[1]{#1}
\csname url@samestyle\endcsname
\providecommand{\newblock}{\relax}
\providecommand{\bibinfo}[2]{#2}
\providecommand{\BIBentrySTDinterwordspacing}{\spaceskip=0pt\relax}
\providecommand{\BIBentryALTinterwordstretchfactor}{4}
\providecommand{\BIBentryALTinterwordspacing}{\spaceskip=\fontdimen2\font plus
\BIBentryALTinterwordstretchfactor\fontdimen3\font minus
  \fontdimen4\font\relax}
\providecommand{\BIBforeignlanguage}[2]{{%
\expandafter\ifx\csname l@#1\endcsname\relax
\typeout{** WARNING: IEEEtran.bst: No hyphenation pattern has been}%
\typeout{** loaded for the language `#1'. Using the pattern for}%
\typeout{** the default language instead.}%
\else
\language=\csname l@#1\endcsname
\fi
#2}}
\providecommand{\BIBdecl}{\relax}
\BIBdecl

\bibitem{mayne2000constrained}
D.~Q. Mayne, J.~B. Rawlings, C.~V. Rao, and P.~O.~M. Scokaert, ``Constrained
  model predictive control: Stability and optimality,'' \emph{Automatica},
  vol.~36, no.~6, pp. 789--814, 2000.

\bibitem{mellinger2011minimum}
D.~Mellinger and V.~Kumar, ``Minimum snap trajectory generation and control for
  quadrotors,'' in \emph{Proc. IEEE International Conference on Robotics and
  Automation}, 2011, pp. 2520--2525.

\bibitem{han2022efficient}
Z.~Han, Y.~Wu, T.~Li, L.~Zhang, L.~Pei, L.~Xu, C.~Li, C.~Ma, C.~Xu, S.~Shen,
  and F.~Gao, ``An efficient spatial-temporal trajectory planner for autonomous
  vehicles in unstructured environments,'' \emph{IEEE Transactions on
  Intelligent Transportation Systems}, vol.~25, no.~2, pp. 1797--1814, 2024.

\bibitem{quan2020survey}
L.~Quan, L.~Han, B.~Zhou, S.~Shen, and F.~Gao, ``Survey of {UAV} motion
  planning,'' \emph{IET Cyber-Systems and Robotics}, vol.~2, no.~1, pp. 14--21,
  2020.

\bibitem{ljung1999system}
L.~Ljung, \emph{System Identification: Theory for the User}, 2nd~ed.\hskip 1em
  plus 0.5em minus 0.4em\relax Upper Saddle River, NJ: Prentice Hall PTR, 1999.

\bibitem{picotti2023gokart}
E.~Picotti, E.~Mion, A.~D. Libera, J.~Pavlovic, A.~Censi, E.~Frazzoli,
  A.~Beghi, and M.~Bruschetta, ``A learning-based nonlinear model predictive
  controller for a real go-kart based on black-box dynamics modeling through
  gaussian processes,'' \emph{IEEE Transactions on Control Systems Technology},
  vol.~31, no.~5, pp. 2055--2065, 2023.

\bibitem{salzmann2023realtime}
T.~Salzmann, E.~Kaufmann, J.~Arrizabalaga, M.~Pavone, D.~Scaramuzza, and
  M.~Ryll, ``Real-time neural {MPC}: Deep learning model predictive control for
  quadrotors and agile robotic platforms,'' \emph{IEEE Robotics and Automation
  Letters}, vol.~8, no.~4, pp. 2397--2404, 2023.

\bibitem{fliess1995flatness}
M.~Fliess, J.~L{\'e}vine, P.~Martin, and P.~Rouchon, ``Flatness and defect of
  non-linear systems: Introductory theory and examples,'' \emph{International
  Journal of Control}, vol.~61, no.~6, pp. 1327--1361, 1995.

\bibitem{richter2016polynomial}
C.~Richter, A.~Bry, and N.~Roy, ``Polynomial trajectory planning for aggressive
  quadrotor flight in dense indoor environments,'' in \emph{Robotics Research},
  ser. Springer Tracts in Advanced Robotics.\hskip 1em plus 0.5em minus
  0.4em\relax Springer, 2016, vol. 114, pp. 649--666.

\bibitem{tal2023aerobatic}
E.~Tal, G.~Ryou, and S.~Karaman, ``Aerobatic trajectory generation for a {VTOL}
  fixed-wing aircraft using differential flatness,'' \emph{IEEE Transactions on
  Robotics}, vol.~39, no.~6, pp. 4805--4819, 2023.

\bibitem{wang2022geometrically}
Z.~Wang, X.~Zhou, C.~Xu, and F.~Gao, ``Geometrically constrained trajectory
  optimization for multicopters,'' \emph{IEEE Transactions on Robotics},
  vol.~38, no.~5, pp. 3259--3278, 2022.

\bibitem{zhou2022swarm}
X.~Zhou, X.~Wen, Z.~Wang, Y.~Gao, H.~Li, Q.~Wang, T.~Yang, H.~Lu, Y.~Cao,
  C.~Xu, and F.~Gao, ``Swarm of micro flying robots in the wild,''
  \emph{Science Robotics}, vol.~7, no.~66, p. eabm5954, 2022.

\bibitem{hewing2020cautious}
L.~Hewing, J.~Kabzan, and M.~N. Zeilinger, ``Cautious model predictive control
  using gaussian process regression,'' \emph{IEEE Transactions on Control
  Systems Technology}, vol.~28, no.~6, pp. 2736--2743, 2020.

\bibitem{viet2024antislossing}
K.~N. Viet, M.~D. Duc, T.~C. Duc, and T.~L. Nguyen, ``Anti-sloshing control:
  Flatness-based trajectory planning and tracking control with an integrated
  extended state observer,'' \emph{IET Cyber-Systems and Robotics}, vol.~6,
  no.~3, p. e12121, 2024.

\bibitem{lu2024aggressive}
G.~Lu, Y.~Cai, N.~Chen, F.~Kong, Y.~Ren, and F.~Zhang, ``Trajectory generation
  and tracking control for aggressive tail-sitter flights,'' \emph{The
  International Journal of Robotics Research}, vol.~43, no.~3, pp. 241--280,
  2024.

\bibitem{lu2025autonomous}
G.~Lu, Y.~Ren, F.~Zhu, H.~Li, R.~Xue, Y.~Cai, X.~Lyu, and F.~Zhang,
  ``Autonomous tail-sitter flights in unknown environments,'' \emph{IEEE
  Transactions on Robotics}, vol.~41, pp. 1098--1117, 2025.

\bibitem{zhou2024disturbanceaware}
X.~Zhou, M.~Wang, C.~Cui, Y.~Wang, C.~Xu, and F.~Gao, ``Internal and external
  disturbances aware motion planning and control for quadrotors,'' \emph{IET
  Cyber-Systems and Robotics}, vol.~6, no.~3, p. e12122, 2024.

\bibitem{zhang2021rapid}
X.~Zhang, M.~Bujarbaruah, and F.~Borrelli, ``Near-optimal rapid {MPC} using
  neural networks: A primal-dual policy learning framework,'' \emph{IEEE
  Transactions on Control Systems Technology}, vol.~29, no.~5, pp. 2102--2114,
  2021.

\bibitem{ma2022identification}
S.~F. Ma, G.~Leylaz, and J.-Q. Sun, ``Identification of differentially flat
  output of underactuated dynamic systems,'' \emph{International Journal of
  Control}, vol.~95, no.~1, pp. 114--125, 2022.

\bibitem{ren2021learning}
H.~Ren, J.~Qi, and P.~Ben-Tzvi, ``Learning flatness-based controller using
  neural networks,'' \emph{ASME Letters in Dynamic Systems and Control},
  vol.~1, no.~2, p. 021003, 2021.

\bibitem{do2025ann}
H.-T. Do, I.~Prodan, and F.~Stoican, ``An {ANN}-enhanced approach for
  flatness-based constrained control of nonlinear systems,'' \emph{Nonlinear
  Dynamics}, vol. 114, no.~7, p. 472, 2026.

\bibitem{yang2025flatness}
F.~Yang, J.~Welde, and N.~Matni, ``Learning flatness-preserving residuals for
  pure-feedback systems,'' in \emph{Proceedings of the 64th IEEE Conference on
  Decision and Control ({CDC})}, 2025, pp. 2439--2446.

\bibitem{hsieh2026formation}
P.-A. Hsieh, F.~Yang, N.~Matni, and M.~A. Hsieh, ``Flatness-preserving residual
  learning for real-time tight quadrotor formation flight,'' \emph{arXiv
  preprint arXiv:2607.12275}, 2026.

\bibitem{zhou2019continuity}
Y.~Zhou, C.~Barnes, J.~Lu, J.~Yang, and H.~Li, ``On the continuity of rotation
  representations in neural networks,'' in \emph{Proc. IEEE/CVF Conference on
  Computer Vision and Pattern Recognition}, 2019, pp. 5745--5753.

\end{thebibliography}

\end{document}